\documentclass[sigconf]{acmart}
\usepackage{enumitem}
\AtBeginDocument{%
  }

\setcopyright{acmlicensed}
\copyrightyear{2018}
\acmYear{2018}
\acmDOI{XXXXXXX.XXXXXXX}
\acmConference[Conference acronym 'XX]{Make sure to enter the correct
  conference title from your rights confirmation email}{June 03--05,
  2018}{Woodstock, NY}
\acmISBN{978-1-4503-XXXX-X/2018/06}

\begin{document}

\title{Target Concept Tuning Improves Extreme Weather Forecasting}
\author{Shijie Ren}
\authornote{Both authors contributed equally to this research.}
\orcid{1234-5678-9012}
\affiliation{%
  \institution{Renmin University of China}
  \city{Beijing}
  \country{China}
}
\email{shj_ren@ruc.edu.cn}

\author{Xinyue Gu}
\authornotemark[1]
\affiliation{%
  \institution{Alibaba Group}
  \city{Hangzhou}
  \country{China}}
\email{guxinyue.gxy@alibaba-inc.com}

\author{Ziheng Peng}
\authornotemark[1]
\affiliation{%
  \institution{Renmin University of China}
  \city{Beijing}
  \country{China}
}
\email{ziheng.peng@ruc.edu.cn}

\author{Haifan Zhang}
\affiliation{%
  \institution{Alibaba Group}
  \city{Hangzhou}
  \country{China}}
\email{zhanghaifan.zhf@alibaba-inc.com}

\author{Peisong Niu}
\affiliation{%
  \institution{Alibaba Group}
  \city{Hangzhou}
  \country{China}}
\email{niupeisong.nps@alibaba-inc.com}

\author{Bo Wu}
\affiliation{%
  \institution{Institute of Atmospheric Physics, Chinese Academy of Sciences}
  \city{Beijing}
  \country{China}
}
\email{wubo@mail.iap.ac.cn}

\author{Xiting Wang}
\authornote{Corresponding author}
\affiliation{%
  \institution{Renmin University of China}
  \city{Beijing}
  \country{China}
}
\email{xitingwang@ruc.edu.cn}

\author{Liang Sun}
\authornotemark[2]
\affiliation{%
  \institution{Alibaba Group}
  \city{Hangzhou}
  \country{China}}
\email{liang.sun@alibaba-inc.com}

\author{Jirong Wen}
\affiliation{%
  \institution{Renmin University of China}
  \city{Beijing}
  \country{China}}
\email{jrwen@ruc.edu.cn}

\renewcommand{\shortauthors}{Reh et al.}

\begin{abstract}

    Deep learning models for meteorological forecasting often fail in rare but high-impact events such as typhoons, where relevant data is scarce. Existing fine-tuning methods typically face a trade-off between overlooking these extreme events and overfitting them at the expense of overall performance. We propose TaCT, an interpretable concept-gated fine-tuning framework that solves the aforementioned issue by selective model improvement: models are adapted specifically for failure cases while preserving performance in common scenarios. To this end, TaCT automatically discovers failure-related internal concepts using Sparse Autoencoders and counterfactual analysis, and updates parameters only when the corresponding concepts are activated, rather than applying uniform adaptation. Experiments show consistent improvements in typhoon forecasting across different regions without degrading other meteorological variables. The identified concepts correspond to physically meaningful circulation patterns, revealing model biases and supporting trustworthy adaptation in scientific forecasting tasks. The code is available at https://anonymous.4open.science/r/Concept-Gated-Fine-tune-62AC.
\end{abstract}

\ccsdesc[500]{Computing methodologies~Knowledge representation and reasoning}

\keywords{Concept-Gated Fine-Tuning, AI Weather Forecasting, Interpretability, Tropical Cyclone Forecasting, Sparse Autoencoders}


\maketitle

\section{Introduction}

Weather forecasting is critical to socio-economic stability. Nearly \$3 trillion in annual private industry revenues, one-third of the sector's total, is exposed to weather and climate risk~\cite{dutton2002opportunities}. In energy management, weather forecasts are vital for predicting load demand and renewable power supply, thereby preventing costly blackouts and curtailment~\cite{hong2020energy}. 

Recently, AI-based forecasting models~\cite{bi2023accurate, lam2023learning} have emerged as powerful alternatives to traditional Numerical Weather Prediction (NWP) systems, achieving high accuracy on standard variables such as temperature and wind with much faster inference~\cite{rasp2024weatherbench,shi2025deep}. However, their performance drops during extreme events such as typhoons (also known as Tropical Cyclones, TC), heatwaves and cold surges. Although rare, these events can have severe consequences, including loss of life, grid collapse, and infrastructure damage. These high-stakes scenarios require reliable forecasts for early warning and emergency response~\cite{katz1997economic}, and reliable deployment requires not only accuracy but also trustworthy models that can diagnose and interpret model behavior when failures occur.\looseness=-1

Achieving such reliability, however, is challenging for extreme weather forecasting due to the severe data imbalance that far exceeds typical imbalanced learning settings. For example, the probability of a typhoon forming within any $5^\circ \times 5^\circ$ region over 24 hours is below 0.039\%~\cite{schumacher2009objective}. Current learning paradigms fail under extreme imbalance because they are inherently a trade-off: both full-model fine-tuning and parameter-efficient methods~\cite{hu2022lora,liu2024dora,zou2023representation,liu2024aligning} mitigate data imbalance by trading off performance across different distributions, rather than correcting model errors in both common and rare scenarios. This trade-off becomes particularly severe in extreme imbalance settings, where hyperparameter tuning becomes highly sensitive~\cite{zhang2022hyperparameter} and resampling or reweighting becomes ineffective~\cite{teng2025self}. Furthermore, the lack of interpretability in these methods limits control over when updates take effect, hinders understanding of why failures occur, and undermines trust—both essential in high-stakes deployments.

We aim to avoid the trade-off between accuracy on rare extreme and common events, while simultaneously providing interpretability during fine-tuning. Our approach is inspired by cognitive neuroscience, which shows that individuals with more modular brain organization exhibit greater neural plasticity when learning new tasks~\cite{gallen2019brain}, suggesting localized task-specific updates minimize interference with previously acquired functions. In contrast, neurons in deep learning models often encode superposed features — multiple concepts mixed within a single activation~\cite{elhage2022toy}— and it is difficult to update specific knowledge without overwriting existing representations~\cite{holton2025parallels}. Motivated by this observation, we propose to fine-tune models within a disentangled representation space, where activations correspond to mono-semantic and interpretable concepts. By targeting only the concepts responsible for failures, our method enables precise, interpretable fine-tuning for extreme weather events without compromising the general ability of the models.\looseness=-1

Specifically, we introduce \textbf{TaCT (Targeted Concept Tuning)}, an interpretable, concept-guided fine-tuning framework for failure-aware model adaptation. TaCT leverages Sparse Autoencoders (SAEs)~\cite{zhang-etal-2023-emergent} to promote functional disentanglement, decomposing superposed representations into quasi-modular, mono-semantic units~\cite{zhang-etal-2023-emergent}. These units naturally correspond to coherent meteorological structures, such as typhoon vortices or high-pressure ridges. Using a small set of extreme-event cases (438 out of 1460 samples), TaCT automatically identifies concepts consistently activated during model failures via continuous counterfactual reasoning. Fine-tuning is then performed in a concept-gated manner, updating newly added model parameters only when failure-associated concepts are active, keeping other predictions unaffected. This mechanism yields concept-level plasticity: similar to that observed in modular human brains, the model learns precisely where and when it fails, without interfering with other functional modules and therefore preserving its general predictive ability.

We demonstrate the effectiveness of this approach on typhoon forecasting across multiple cyclone basins (Northern Atlantic, Western Pacific and Eastern Pacific). The resulting models achieve prominent improvements across key typhoon characteristics including sea-level pressure (9.3\% MAE reduction in 72h forecast) and near-surface winds (4.8\% MAE reduction in 72h forecast), while maintaining the forecast accuracy on other variables (error change: ours -2 vs LoRA +4 in z850 and 0 vs +0.1 in T850). Beyond performance gains, the concepts identified by TaCT correspond to physically meaningful atmospheric structures, such as mid-latitude transient waves, which meteorologists recognize as key drivers of typhoon behavior. TaCT not only improves typhoon forecast performance but also reveals concepts that have been improved during fine-tuning, fostering trust and facilitating integration into operational weather prediction workflows.

Our contributions are as follows:

\begin{itemize}[nosep,leftmargin=1em,labelwidth=*,align=left]
    \item We propose TaCT, an interpretable fine-tuning framework that leverages brain-inspired modular learning to disentangle superposed representations into physically grounded concepts, and uses them to guide targeted adaptation. It serves as a generic add-on for intermediate layers and is also generalizable for other deep learning domains.
    
    \item We introduce a counterfactual concept localization method, which integrates SAE with continuous counterfactual reasoning to automatically identify concepts to be improved in extreme weather conditions. This module requires only a small amount of extreme weather data points and can operate fully automatically without manual intervention.  

\item We design a concept-gated fine-tuning algorithm that conditions parameter updates on the activation of failure-related concepts. This module precisely corrects extreme-event failures while preserving performance in common weather regimes, avoiding the trade-off between rare-event performance and overall accuracy.

\end{itemize}

\section{Problem Formulation}
We consider the standard AI weather forecasting task~\cite{chen2023fuxi}: given the current state of the weather (e.g., temperature, humidity, wind speed), predict its future state at a specified lead time. 
The input - global weather data - is represented as a 3D tensor discretized over a latitude-longitude grid. To reduce memory consumption, the input is typically partitioned into patches, each represented as a single unit for model input. Formally, let $X_t \in \mathbb{R}^{H \times W \times C}$ denote the weather conditions at time $t$, where $C$ is the number of weather variables, and $H$ and $W$ denote the number of patches along latitude and longitude, respectively.
Our goal is to predict 
$X_{t+\Delta t} \in \mathbb{R}^{H \times W \times C}$ at lead time 
$t+\Delta t$ using an AI weather model $\mathcal{F}$ with parameters $\Theta$:\looseness=-1
\begin{equation}
\hat{X}_{t+\Delta t} = \mathcal{F}(X_t;\Theta)
\end{equation}

For long-range weather forecasting, predictions are made autoregressively. Specifically, to predict weather conditions at time $t+a\Delta t$, we iteratively apply the model with lead time $\Delta t$ for $a$ times, feeding each prediction back as input for the next step:
\begin{equation}
\label{equ:multi_forward}
\hat{X}_{t+a\Delta t}  = \mathcal{F}^{(a)}(X_t; \Theta) = \underbrace{\mathcal{F}(\cdots\mathcal{F}(\mathcal{F}}_{a\text{ times}}(X_t; \Theta); \Theta)\cdots; \Theta)
\end{equation}
where $\mathcal{F}^{(a)}$ denotes $a$ iterative applications of $\mathcal{F}$.

\section{Methods}
\begin{figure*}[ht] 
    \centering
    \includegraphics[width=0.85\textwidth]{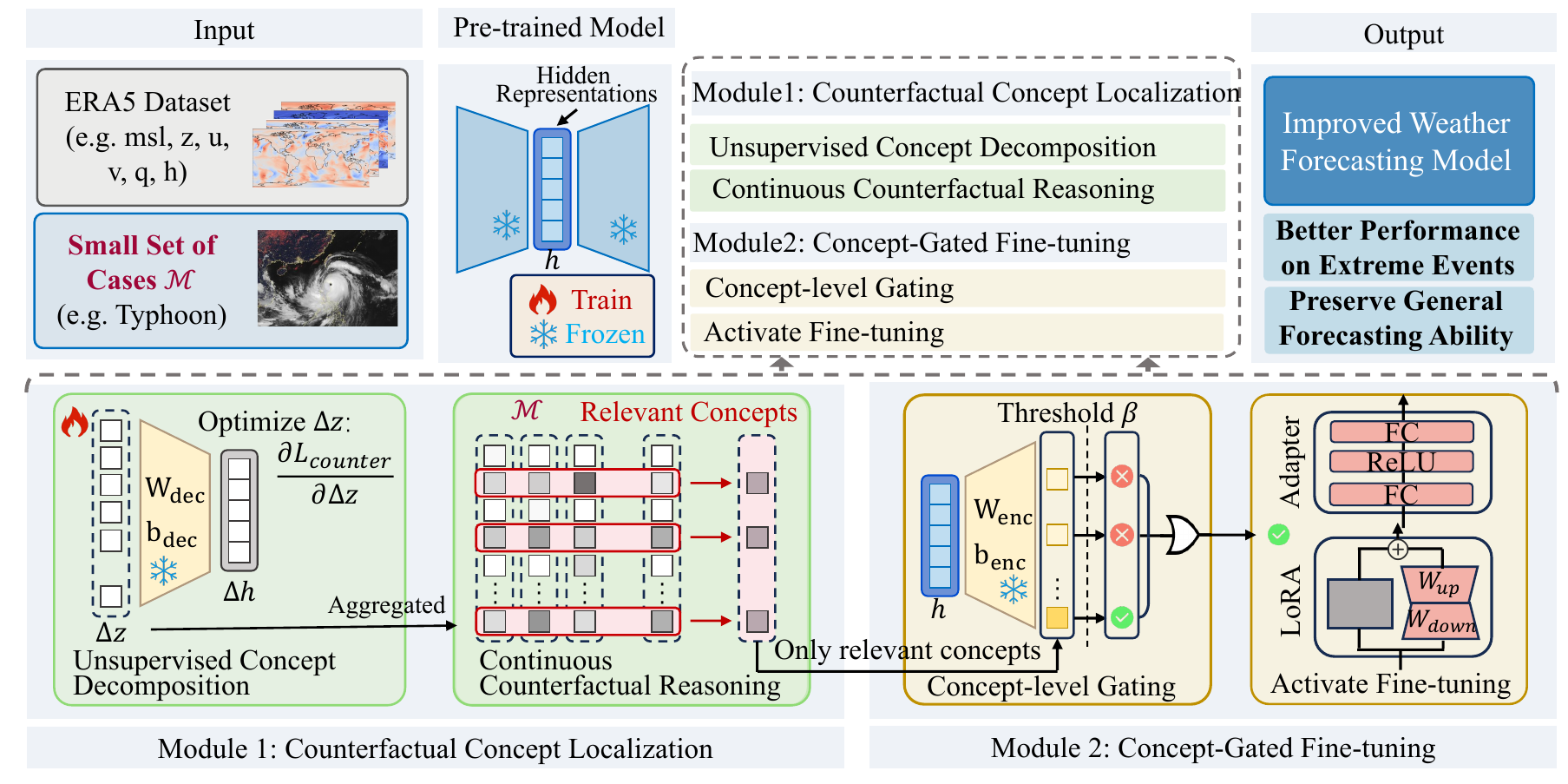} 
    \Description{The overall framework of TaCT}
    \caption{The overall framework of TaCT, comprising two main modules: 
(1) counterfactual concept localization, which decomposes input hidden representations 
and identifies key concepts through
continuous counterfactual reasoning; (2) concept-gated fine-tuning, which 
selectively fine-tunes activated concepts without affecting others.} 
    \label{Fig:method} 
\end{figure*}

This section introduces the  Targeted Concept Tuning (TaCT) framework, which enables interpretable and precise fine-tuning for extreme weather events
without compromising the general abilities.
Analogous to modular human brains~\cite{gallen2019brain}, TaCT fine-tunes in a disentangled representation space, where each activation corresponds to a mono-semantic, quasi-modular concept, enabling targeted adaptation without overwriting unrelated knowledge.

Specifically, our framework consists of two major modules as shown in Figure~\ref{Fig:method}: 
\textbf{(1) counterfactual concept localization}, which disentangles the model's internal representations into mono-semantic concepts without any supervision signals about which concepts exist for the given model, and then identifies key concepts relevant to specific weather scenarios (e.g., extreme events or regional biases) via counterfactual analysis; and
\textbf{(2) concept-gated fine-tuning}, which performs selective updates to model parameters based on the localized target concepts.



Next, we will introduce the two modules and discuss the theoretical relation between TaCT and existing fine-tuning methods. 

\subsection{Counterfactual Concept Localization}

This module disentangles the model’s hidden representations into interpretable concepts and identify those most responsible for prediction failures under the target scenario (i.e., extreme weather scenarios). Given a small set of extreme weather samples $(X_t, X_{t+\Delta t})$, we obtain a concept set $C$ that guides the subsequent fine-tuning through two stages: unsupervised concept decomposition and continuous counterfactual reasoning.

\subsubsection{Unsupervised Concept Decomposition}
Given hidden embedding $H_l = \mathcal{F}_{1:l}(X_t) \in \mathbb{R}^{H \times W \times d}$ from layer $l$, the goal of unsupervised concept decomposition is to decompose $H_l = \mathcal{F}_{1:l}(X_t)$ into sparse concepts $Z \in \mathbb{R}^{H \times W \times n}$ ($n>d$) without additional supervision signal about which concepts exist in the weather forecasting models.  We achieve this goal by employing Sparse Autoencoders (SAEs)~\cite{bussmann2024batchtopk,gaoscaling}, which constructs a disentangled concept space $Z$ by learning sparse representations with an auto-encoder while accurately reconstructing the original input. Each concept $z^{(i,j)}$ in $Z$ then corresponds to a mono-semantic feature relevant to meteorological structures (e.g., typhoon vortices or pressure ridges).


Mathematically, given a sample $X^{(i,j)}$ at geographic coordinate $(i,j)$, the encoder projects the $l$-th layer hidden representation $h_{l}^{(i,j)}\in \mathbb{R}^{d}$ into a high-dimensional sparse concepts space $z^{i,j} \in \mathbb{R}^n$ ($n > d$), and the decoder reconstructs the original embedding:
\begin{equation}
    \label{equ:enc}
    z^{(i,j)} = \text{Enc}(h_{l}^{(i,j)}) = \text{Topk}(\text{ReLU}(W_{enc}(h_{l}^{(i,j)} - b_{dec}) + b_{enc},k)),
\end{equation}
\begin{equation}
     \hat{h}_{l}^{(i,j)} = \text{Dec}(z^{(i,j)}) = W_{dec}z^{(i,j)} + b_{dec},
\end{equation}
where $W_{enc} \in \mathbb{R}^{n \times d}$,  $W_{dec} \in \mathbb{R}^{d \times n}$, $b_{enc} \in \mathbb{R}^{n}$, $ b_{dec}\in  \mathbb{R}^{d} $ are parameters to be learned, and TopK$(\cdot)$ retains only the $k$ largest activations and setting other elements to 0, ensuring sparse activation.


The encoder-decoder is then learned to reconstruct the hidden representations accurately, with an auxiliary loss function $Aux(\cdot)$~\cite{gaoscaling} to reactivate dead concepts whose activations $z^{(i,j)}$ are consistently low for almost all samples:
\begin{equation}
    \label{equ:SAE_loss}
    \mathcal{L}_{SAE} = | \hat{h}_{l}^{(i,j)} - h_{l}^{(i,j)} |_2^2 + \lambda Aux(z^{(i,j)}).
\end{equation} 
Appendix~\ref{app:sae_details} introduces $Aux(\cdot)$ and SAE training in detail.

\subsubsection{Continuous Counterfactual Reasoning}
\label{sec:Counterfactual}
Given $n$ discovered concepts, where $n>d$ is usually large, we then discover a subset $C$ of concepts that is most responsible for prediction failures under extreme weather scenarios. This requires quantifying each concept’s contribution to the forecasting loss. Our idea is based on counterfactual reasoning~\cite{abid2022meaningfully,ross2021explaining,pan2024counterfactual}, which estimates concept importance by measuring how minimal interventions on individual concepts affect losses in a small set of extreme weather samples:
\begin{equation}
    \label{equ:conterfactual}
    \min_{z'} \text{dist}(z, z') \quad \text{s.t.} \quad  F_{l+1:L}(\text{Dec}(z')) = y_{target}.
\end{equation}
Here, we omit the position index $(i,j)$ of $z$ for conciseness, and the minimal intervention magnitude $|z- z'|$ represents the importance of the corresponding concepts in decreasing loss.

Unlike prior counterfactual methods shown in the last equation, which are designed for discrete classification, weather forecasting is a continuous regression problem. We therefore change the hard constraint $F_{l+1:L}(\text{Dec}(z')) = y_{target}$ into a soft reconstruction objective and optimize perturbations directly in concept space, enabling continuous counterfactual reasoning:

\begin{equation} 
\label{equ:conterfac}
\min_{\Delta_{z}} \mathcal{L} = |F_{l+1:L}(\text{Dec}(z'))-y_{target}|_2^2 + \lambda |z'-z|,
\end{equation}
\begin{equation}
    y_{target}=X_{t+\Delta_t}^{(i,j)}, \text{    } z'=z + \mathbb{I}(z>0)\Delta{z},
\end{equation}
\begin{equation}
    \Delta z=z'-z.
\end{equation}

Here, the $L_1$ penalty encourages sparse perturbations, $ \mathbb{I}(\cdot)$ is an indicator function where $\mathbb{I}(\text{true})=1$ and $\mathbb{I}(\text{false})=0$. $z'=z + \mathbb{I}(z>0)\Delta{z}$ ensures that a counterfactual intervention will be conducted only for relevant samples. Intuitively, each entry in $\Delta z=z'-z$ measures how much a specific concept must change to reduce prediction error: larger magnitudes indicate concepts whose adjustment most improves the forecast.

Finally, the top-k concepts with the highest average magnitudes across all $m$ extreme weather samples are then selected as the target concepts for subsequent modification. This selection process is formally defined in Eq.~\eqref{equ:concept-top}:

\begin{equation} 
\label{equ:concept-top}
\mathcal C=\text{Topk}_{id}(\frac{1}{m} \sum_{t=1}^{m} |\Delta{z}_{t}|,k).
\end{equation}
where the resulting set $\mathcal C$ contains the key concepts to be fine-tuned for the target scenario. These concepts exhibit the most significant influence on the model's performance for the target task, and $\text{Topk}_{id}$ is used to find the indices of the $k$ largest elements.

\subsection{Concept-Gated Fine-tuning}
To enhance specific model capabilities by modifying the concepts identified through counterfactual reasoning—while simultaneously preserving its general performance—we introduce a concept-based gating mechanism. Specifically, this mechanism leverages the activation states of concepts to precisely control whether the current data point should be used to optimize a particular concept. Such a structured fine-tuning strategy ensures that biases are corrected with precision, thereby minimizing any degradation of the model's performance on other data.

\subsubsection{Concept-Gated Fine-tuning}
Previous work has demonstrated that certain knowledge can be modified or forgotten by manipulating specific neurons~\cite{dai2022knowledge}. Inspired by this, we introduce a concept activation-gated mechanism to precisely modify performance under specific concepts without affecting the model's performance in other scenarios, such as extreme low pressure in typhoon scenarios. Specifically, for each concept $c_i$ in the concept set $\mathcal{C}$ identified in Section~\ref{sec:Counterfactual}, we define an threshold $\beta_i$. This threshold ensures that the fine-tuned parameters are only activated when the corresponding concept is triggered. Formally, for the $l$-th layer targeted for intervention, let the intermediate representation be $\mathbf{h}_l$. The concept activation vector obtained through SAE encoding is denoted as $\mathbf{z} = \text{Enc}(\mathbf{h}_l)$. We modify the model's forward pass by introducing a gated residual term. This controlled modification term, $\Delta \mathbf{h}_l$, is computed as shown in Eq.~\eqref{equ:gate}, where $\mathbb{I}(\cdot)$ is the indicator function shown in Eq.~\eqref{equ:indicator}, which determines whether the activation of the $i$-th concept exceeds the threshold $\beta_i$. Here, $f'$ represents an existing fine-tuning method, such as LoRA or Adapter.

\begin{equation}
    \label{equ:gate}
    \Delta \mathbf{h}_l =  \mathbb{I}\left(\vee_{i \in C} (z_i > \beta_i)\right) \cdot f'(\mathbf{h}_l).
\end{equation}

\begin{equation}
\label{equ:indicator}
\mathbb{I}(\vee_{i \in C}~(z_i > \beta_i)) = 
\begin{cases} 
1, & \text{if Ture}  \\
0, & \text{otherwise} 

\end{cases}.
\end{equation}

When the input does not involve the target concepts, the term $\Delta \mathbf{h}_l$ vanishes as the indicator function evaluates to zero. Finally, we inject the generated gated residual into the original model to obtain the modified final prediction, formalized as Eq.~\eqref{equ:intervene}, where $F_{l+1:L}$ represents the subsequent processing of the model from the intervention layer to the output layer, $\hat{X}'_{t+\Delta{t}}$ represents the model output after intervention. 

\begin{equation}
    \label{equ:intervene}
    \hat{X}'_{t+\Delta{t}} = F_{l+1:L}(\mathbf{h}_l + \Delta \mathbf{h}_l).
\end{equation}

\subsubsection{Loss Function}

The loss function follows previous work~\cite{chen2023fuxi}, employing a latitude-weighted loss, formulated as Eq.~\eqref{equ:loss}, where $H$, and $W$ denote the number of grid points in the latitude and longitude directions, respectively. $\hat{X}'$ and $X$ are the predicted and ground-truth values for a specific variable and location (latitude and longitude) at time step $t+\Delta{t}$. $w_i$ represents the weight at latitude $i$, which decreases as the latitude increases. The loss is averaged across all grid points and variables.
\begin{equation}
    \label{equ:loss}
    \mathcal{L} = \frac{1}{HW}\sum_{i=1}^{H} \sum_{j=1}^{W} w_i \cdot | \hat{X}'_{t+\Delta{t}} - X_{t+\Delta{t}}|.
\end{equation}

\subsection{Method Comparison}
\label{sec:comp}
From the perspective of concept intervention, different model editing methods essentially adopt different strategies for intervening on concepts. Our method proposes a more general framework for concept-gated intervention.

``Black-box'' methods, such as PEFT, employ a global and indiscriminate intervention strategy. They do not differentiate based on concept importance, applying adjustments uniformly across all internal states. This can be viewed as a special case with the most lenient intervention condition: the gating unit is always open, intervening on all concepts $C_{\text{all}}$, as formalized in Eq.~\eqref{eq:peft_intervention}.
\begin{equation} \label{eq:peft_intervention}
    \mathbb{I}_{\text{LoRA}} = \mathbb{I}(\vee i \in C_{\text{all}} (z_i > 0)) = 1 \quad.
\end{equation}

Representation engineering is another approach where intervention concepts are pre-defined. The intervention target $C_{\text{pre}}$, relies entirely on human prior knowledge and manual specification. An intervention is executed only when a pre-defined concept is activated, as formalized in Eq.~\eqref{eq:repe_intervention}.
\begin{equation} \label{eq:repe_intervention}
    \mathbb{I}_{\text{RepE}} (\vee i \in C_{\text{pre}} (z_i > \beta^{\text{pre}}_i)) =
    \begin{cases}
        1, & \text{if True} \\
        0, & \text{otherwise}
    \end{cases}.
\end{equation}

Unlike PEFT’s indiscriminate updates and ReFT’s reliance on pre-defined concepts. Our methods learns to automatically discover when and on which key concepts to intervene, thereby enabling automated form of concept intervention.


\section{Experiments}
\subsection{Experimental Settings}
\textbf{Datasets}. We utilize the ERA5 dataset~\cite{hersbach2020era5} for model training and test as both inputs and ground truths. Produced by the European Centre for Medium-Range Weather Forecasts (ECMWF), ERA5 is a comprehensive global atmospheric reanalysis dataset providing detailed climate and  weather information from 1940 to the present. It encompasses a wide array of variables, including temperature, humidity, precipitation, and mean sea level pressure. ERA5 offers a high spatial resolution of 0.25° latitude-longitude and includes 37 vertical pressure levels. We adopt the International Best Track Archive for Climate Stewardship (IBTrACS)~\cite{knapp2010international} and CMA Best Track Dataset~\cite{ying2014overview,lu2021western} as our primary data source. IBTrACS is a global typhoon dataset maintained by the U.S. National Oceanic and Atmospheric Administration (NOAA) and is widely considered the gold standard for research in this field. The CMA Best Track Dataset is a typhoon best track dataset released and maintained by the Shanghai Typhoon Institute (STI) under the China Meteorological Administration (CMA). It is one of the most authoritative historical records of typhoon over the Western North Pacific and the South China Sea.

\noindent\textbf{Foundation Model}. The foundation weather forecasting model used in this work is Baguan~\cite{niu2025utilizing}, a large-scale pre-trained weather model developed by Alibaba. Baguan is a deep learning model designed for global medium-range weather forecasting, trained end-to-end on the ERA5 dataset with robust spatiotemporal modeling capabilities. It operates at a spatial resolution of 0.25° × 0.25° and covers major meteorological factors such as geopotential, wind speed, temperature, humidity, and precipitation, with a 6-hour time step. Benefiting from extensive pre-training on large-scale weather data and the implicit learning of atmospheric physics, Baguan has demonstrated predictive accuracy comparable to traditional Numerical Weather Prediction (NWP) systems.

\noindent\textbf{Baseline and Metric}. To provide a comprehensive evaluation, we compare with four baselines: the Baguan-Weather forecasting model, two parameter-efficient fine-tuning (PEFT) methods: LoRA and Adapter and a representation-based fine-tuning method: LoREFT. The implementation details and hardware configuration can be found in the Appendix~\ref{app:details}. Our evaluation metrics are the minimum mean sea-level pressure (MSL) and the maximum 10-meter wind speed (V10) of tropical cyclones.

\noindent\textbf{Implementation Details}. To evaluate the model's performance, typhoon data from 2022 is used exclusively as an external test set to ensure complete isolation from the training process and prevent potential data leakage. The training phase utilizes unlabeled ERA5 data spanning from 1979 to 2021. Due to limited computational resources, we restricted the optimization process by uniformly fixing the number of training steps at 6,000. This approach ensures model convergence while maximizing overall training efficiency.

\subsection{Overall Performance}
\begin{figure} 
    \centering
    \includegraphics[width=0.48\textwidth]{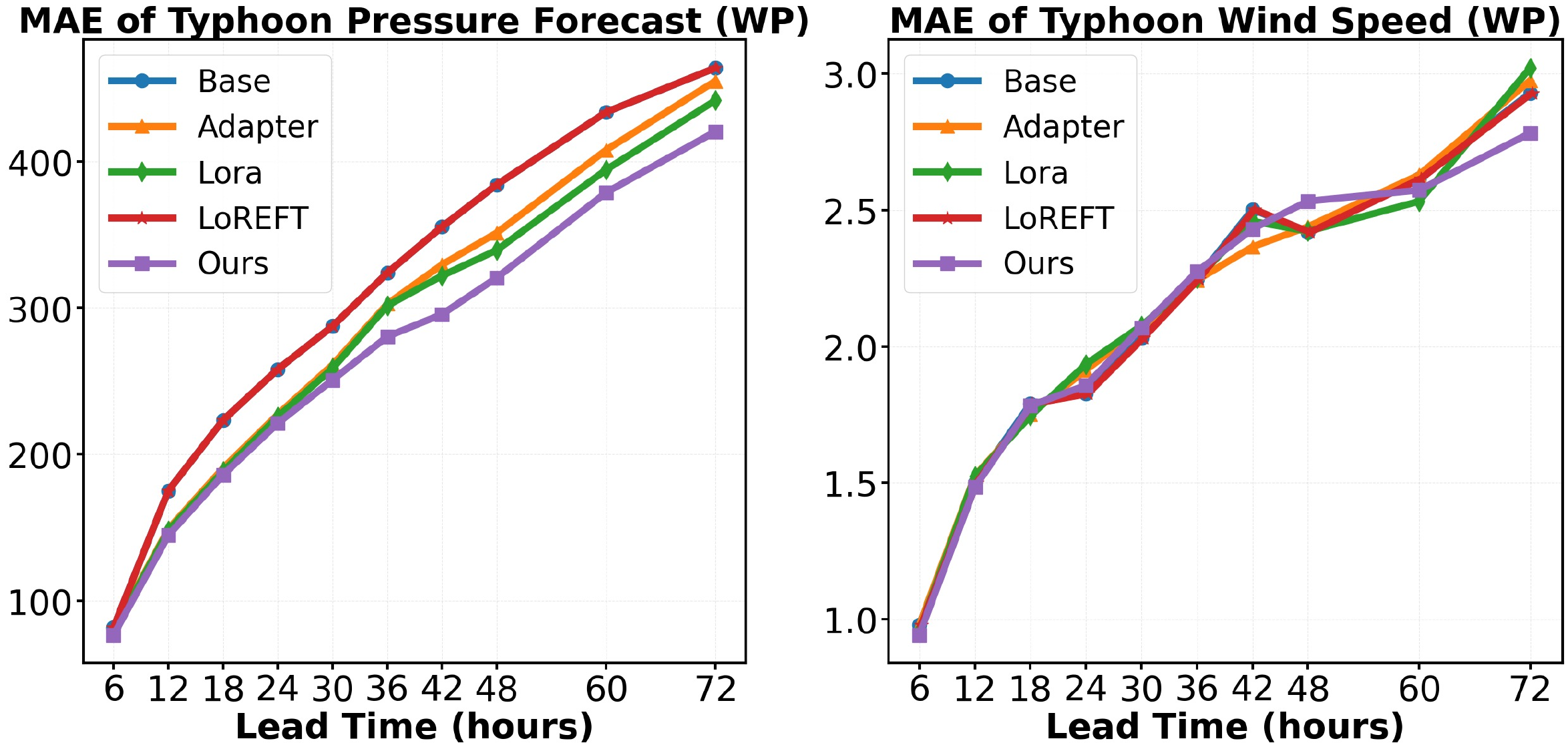} 
    \caption{72-Hours Performance. WP denotes the Western Pacific. “Typhoon pressure” refers to the minimum sea-level pressure in Pa, and “Typhoon wind speed” represents the maximum wind speed in m/s.} 
    \label{Fig:performance} 
\end{figure}

\noindent\textbf{Results}. We evaluated the 72-hour forecast accuracy of typhoon intensity, including MSL and V10, across multiple regions. The results for the Western Pacific are illustrated in Figure~\ref{Fig:performance}, while the remaining results are presented in Appendix~\ref{app:72}. TaCT demonstrated comparable performance for both typhoon minimum pressure and wind speed forecasts. The performance of LoREFT is highly similar to that of the Base Model. This is because LoREFT only modifies a subset of tokens, which may be located far from the typhoon center. As AI weather models typically focus on data within a local neighborhood~\cite{niu2025utilizing}, the impact of these distant modifications is minimal. In contrast, Lora and Adapter achieved sub-optimal results due to a lack of fine-grained conceptual guidance. To compare with models fine-tuned on tropical cyclone data, we evaluate our method against several fine-tuning approaches trained on the same tropical cyclone dataset. The experimental results are presented in Appendix~\ref{app:typhoon}.

\begin{figure}[ht] 
    \centering
    \includegraphics[width=0.48\textwidth]{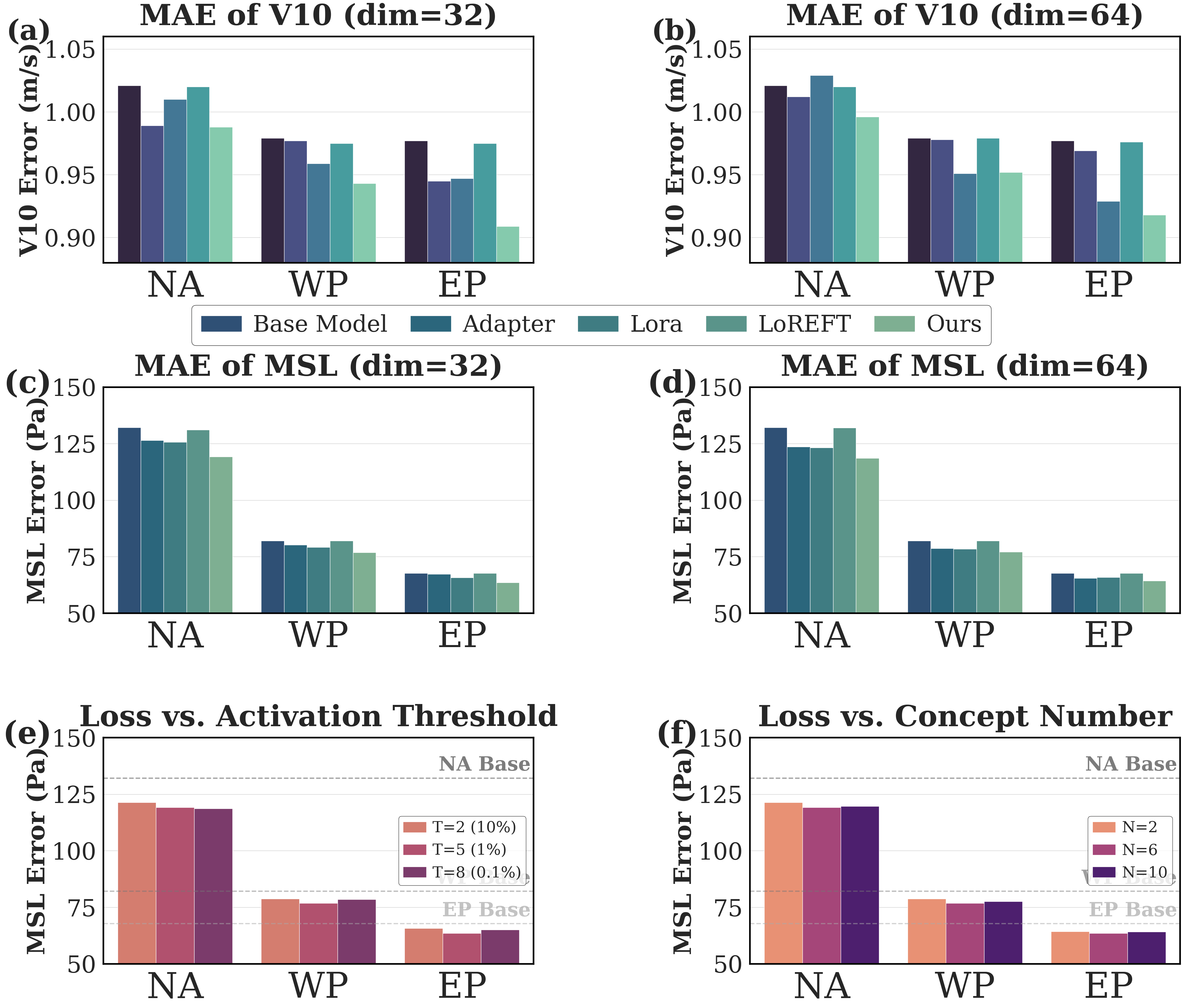} 
    \caption{Sensitivity Analysis. (a, b) and (c,d) present the 6-hour forecast results of minimum sea level pressure (MSL) and maximum wind speed of typhoons across different regions under varying parameter sizes. (e) presents the 6-hour MSL forecast results under different model activation threshold settings. (f) presents the MSL forecast results under different concept number settings.} 
    \label{Fig:sensitivity} 
\end{figure}

\subsection{Ablation Study and Sensitivity Analysis}
\textbf{Ablation Study}. We conduct an ablation study to evaluate the contribution of each component in our framework. To assess the effect of the concept-gated module, we remove the concept-gated unit, which degrades the method to a simple Adapter. To evaluate the continuous counterfactual reasoning module, we consider two ablated variants: one with randomly sampled concepts and another using a statistical method for concept identification. The specific implementation details are provided in Appendix~\ref{app:statistical}. The results in Table~\ref{tab:ablation} demonstrate that the interpretability module plays the primary role in enhancing model performance. Replacing our components with these alternatives leads to a consistent drop in performance, thereby validating the effectiveness of our proposed method.

\noindent \textbf{Threshold of Concepts}. To evaluate the impact of the number of gated concepts on model performance. We choose three activation thresholds that filter out 90\%, 99\%, and 99.9\% of the data, respectively, as shown in Figure~\ref{Fig:sensitivity}~(e). The results clearly indicate that the number of concepts is correlated with model effectiveness. Within a certain range, increasing the number of concepts leads to a lower MAE, as the model tends to learn better representations when more concepts are adjusted. However, once the number of concepts reaches a certain threshold, model performance begins to decline. This is because too many concepts can introduce more noise, degrading the quality of the learned concepts.

\noindent \textbf{Number of Concepts}. We evaluate the effect of the number of gated concepts on model performance, as shown in Figure~\ref{Fig:sensitivity}~(f). The results clearly indicate that the number of concepts is correlated with model effectiveness. Within a certain range, increasing the number of concepts leads to a lower MAE, as the model tends to learn better representations with more concepts being adjusted. However, once the number of concepts reaches a certain threshold, model performance begins to decline. This is because a large number of concepts can introduce more noise, leading to a degradation in the quality of the learned concepts. 

\noindent \textbf{Randomized Scenarios}. 
To minimize randomness as much as possible, we evaluated our method with three different random seeds and under varying fine-tuning parameter budgets, as shown in Figure~\ref{Fig:sensitivity}~(a-d) and Table~\ref{tab:seed}. The results demonstrate that our method consistently outperforms the baseline across different settings.


\begin{table}[t]
  \centering
  \caption{Ablation study: MAE of MSL under different methods. M1 denotes counterfactual concept localization, and ``random'' and ``statistic'' refer to two different concept localization strategies.}
  \label{tab:ablation}
  \small
  \setlength{\tabcolsep}{6pt}
  \begin{tabular}{lcccc}
    \toprule
    \textbf{Region} & \textbf{w/o M1} & \textbf{Random} & \textbf{Statistics} & \textbf{Ours} \\
    \midrule
    West Pacific   &80.21& 82.30 & 78.09 & \textbf{76.88} \\
    East Pacific   &67.34& 67.97 & 64.80 &  \textbf{63.53} \\
    North Atlantic &126.50& 131.48 & 121.36 & \textbf{119.21} \\
    \midrule
    \textbf{Avg}  &91.35 &93.92 &88.08 &\textbf{86.54} \\
    \textbf{$\Delta$ (\%)} & 5.27 & 7.85 & 1.75 & - \\
    \bottomrule
  \end{tabular}
\end{table}

\begin{table}[t]
  \centering
  \caption{MAE of MSL under different random seeds.}
  \label{tab:seed}
  \small
  \setlength{\tabcolsep}{8pt}
  \begin{tabular}{lccc}
    \toprule
    \textbf{Region} & \textbf{Seed=1979} & \textbf{Seed=2025} & \textbf{Seed=2026} \\
    \midrule
    West Pacific   & 76.88  & 77.92  & 75.84 \\
    East Pacific   & 63.53  & 64.06  & 61.97 \\
    North Atlantic & 119.21 & 121.55 & 116.56 \\
    \bottomrule
  \end{tabular}
\end{table}



\subsection{General Capabilities}
\begin{figure*}[t] 
    \centering
    \includegraphics[width=1\textwidth]{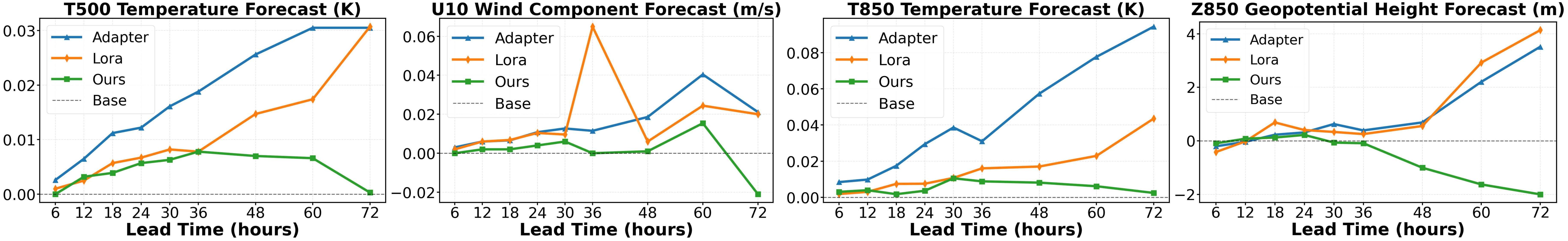} 
    \caption{The impact of different fine-tuning methods on general capabilities. Where 0 represents the performance of the base model, and the values indicate the change of different methods relative to the original model, with lower values being better.} 
    \label{exp:general} 
\end{figure*}

To verify the impact of fine-tuning on the model's general capabilities, we conducted a comprehensive evaluation of the model's performance on global weather forecasting tasks. We validate the model's 72-hour forecast performance across the entire globe. The performance changes relative to the base model are illustrated in Figure~\ref{exp:general}, where values closer to zero indicate better preservation of general capabilities. The experimental results demonstrate that our method not only achieves significant performance improvements in typhoon-specific scenarios but also maintains the strongest general forecasting capabilities among all compared fine-tuning approaches. Notably, our approach exhibits minimal degradation across all evaluated variables, with performance changes consistently smaller than those of other methods, confirming the effectiveness of our counterfactual concept localization strategy in balancing task-specific optimization with the retention of foundational model knowledge.

\subsection{Case Study}
To interpret the physical meaning of these concepts, we adopt an automated approach based on Multimodal Large Language Models (MLLMs). In brief, the MLLM-based method feeds a concept’s activation locations together with its corresponding input–output information into an MLLM, leveraging its strong pre-trained capabilities to summarize the concept’s function. More details are provided in Appendix~\ref{app:inter}.

\subsubsection{Concepts-general Analysis}
\begin{figure}[ht]
    \centering
    \includegraphics[width=0.48\textwidth]{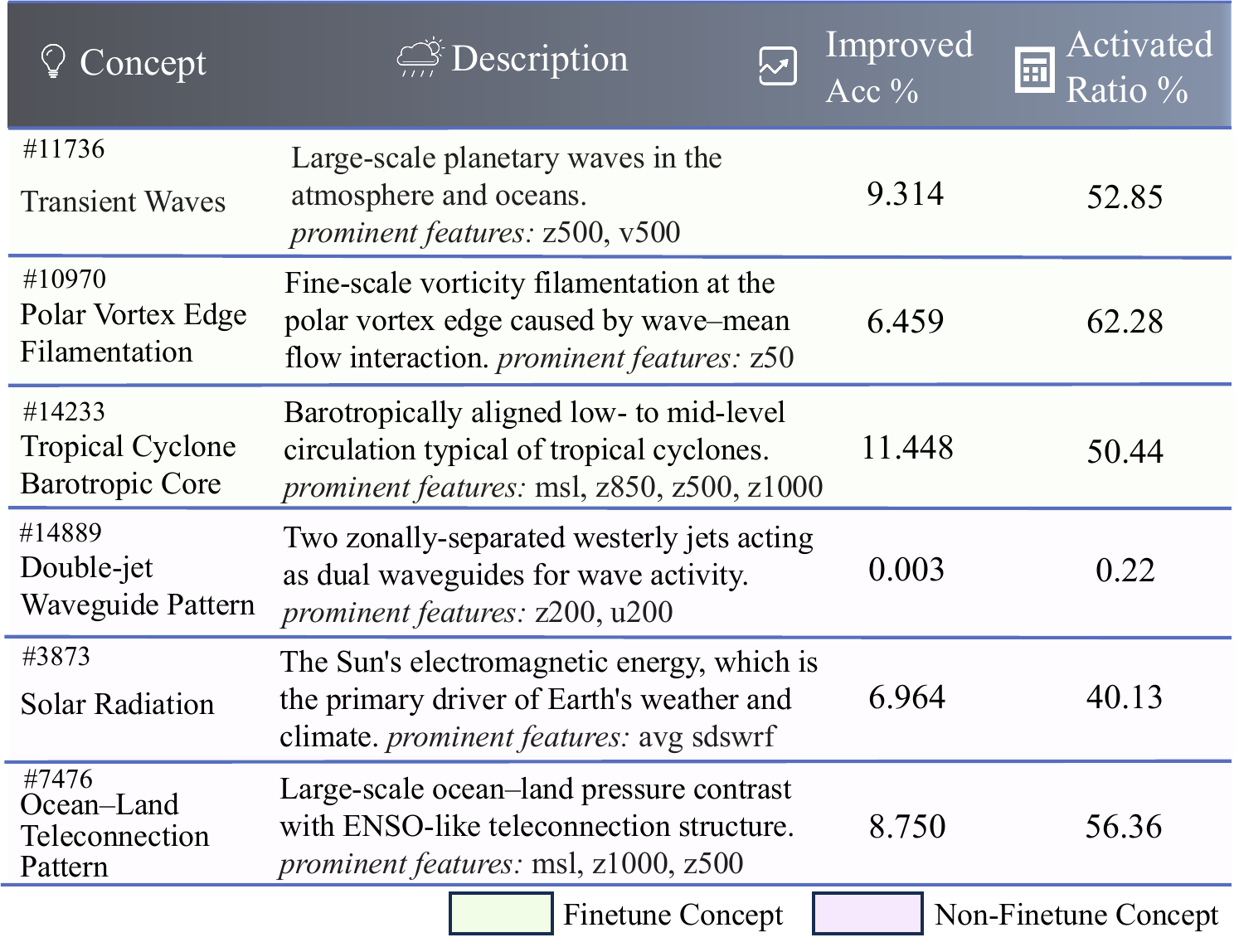} 
    \caption{Six concepts learned by SAE.}
    \label{exp:conceptlist}
\end{figure}

To gain a deeper understanding of the physical concepts learned by the model, we first analyze the list of concepts presented in Figure~\ref{exp:conceptlist}. This table enumerates a series of concepts, including their names and detailed descriptions as generated by a MLLM. The \textit{Improved Acc} column indicates the proportion in Mean Absolute Error (MAE) after fine-tuning when a concept is activated; a higher value signifies a more substantial improvement in model performance. The \textit{Activate Ratio} column quantifies the number of typhoon cases in which the concept was activated.

From the data, several key observations can be made. The first three concepts, namely \textbf{\#11736}, \textbf{\#10970}, and \textbf{\#14233}, which were identified by our counterfactual concept localization method, are activated in a large number of typhoon cases. They also significantly improve the model's predictive accuracy. This demonstrates that the model successfully captures critical physical patterns closely associated with typhoon events.

In contrast, the remaining three concepts, namely \textbf{\#14889}, \textbf{\#3873}, and \textbf{\#7476}, were not identified by our counterfactual concept localization method. Concept \textbf{\#14889} is rarely activated in typhoon cases (only once), suggesting its correlation with the typhoon phenomenon is weak. It is also noteworthy that concepts \textbf{\#3873} and \textbf{\#7476}, although not directly identified by counterfactual concept localization, are also relevant to typhoon phenomena. Consequently, they are activated in many cases and contribute positively to the model's performance enhancement.

These findings validate the effectiveness of our proposed method in automatically identifying and utilizing physical concepts related to specific extreme weather events like typhoons, thereby laying the groundwork for the subsequent case study analysis.

\subsubsection{Concept-specific Cases}
We further illustrate a critical concept 11736 activations whose semantic meanings are interpreted by an MLLM. 

\begin{figure}[ht] 
    \centering
    \includegraphics[width=0.5\textwidth]{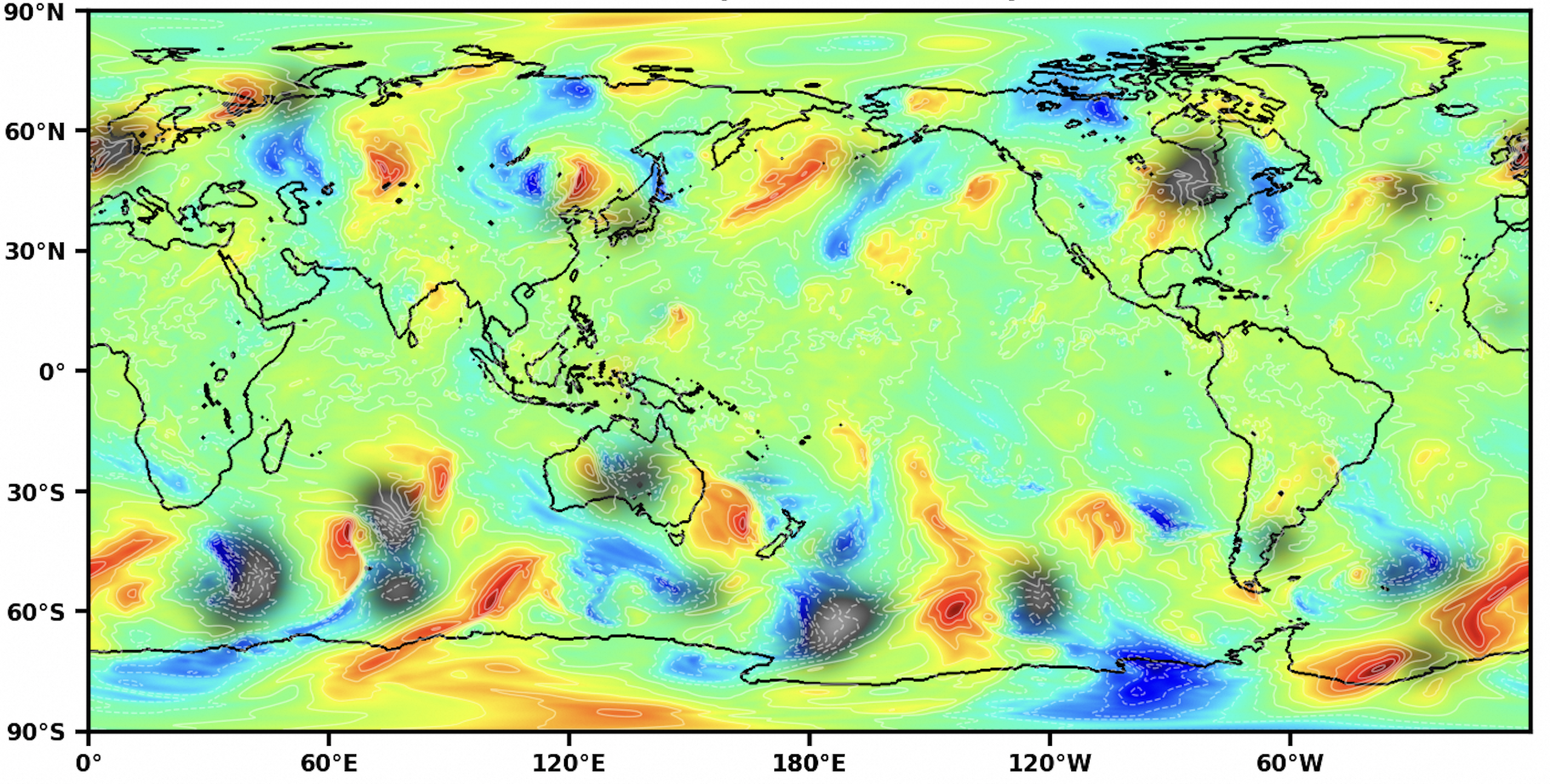} 
    \caption{V-component of wind on 500hPa, where red indicates positive values, 
blue indicates negative values, and black mask indicates concept 
activation locations.} 
        \label{Fig:v500}
\end{figure}

\begin{figure}[ht] 
    \centering
    \includegraphics[width=0.5\textwidth]{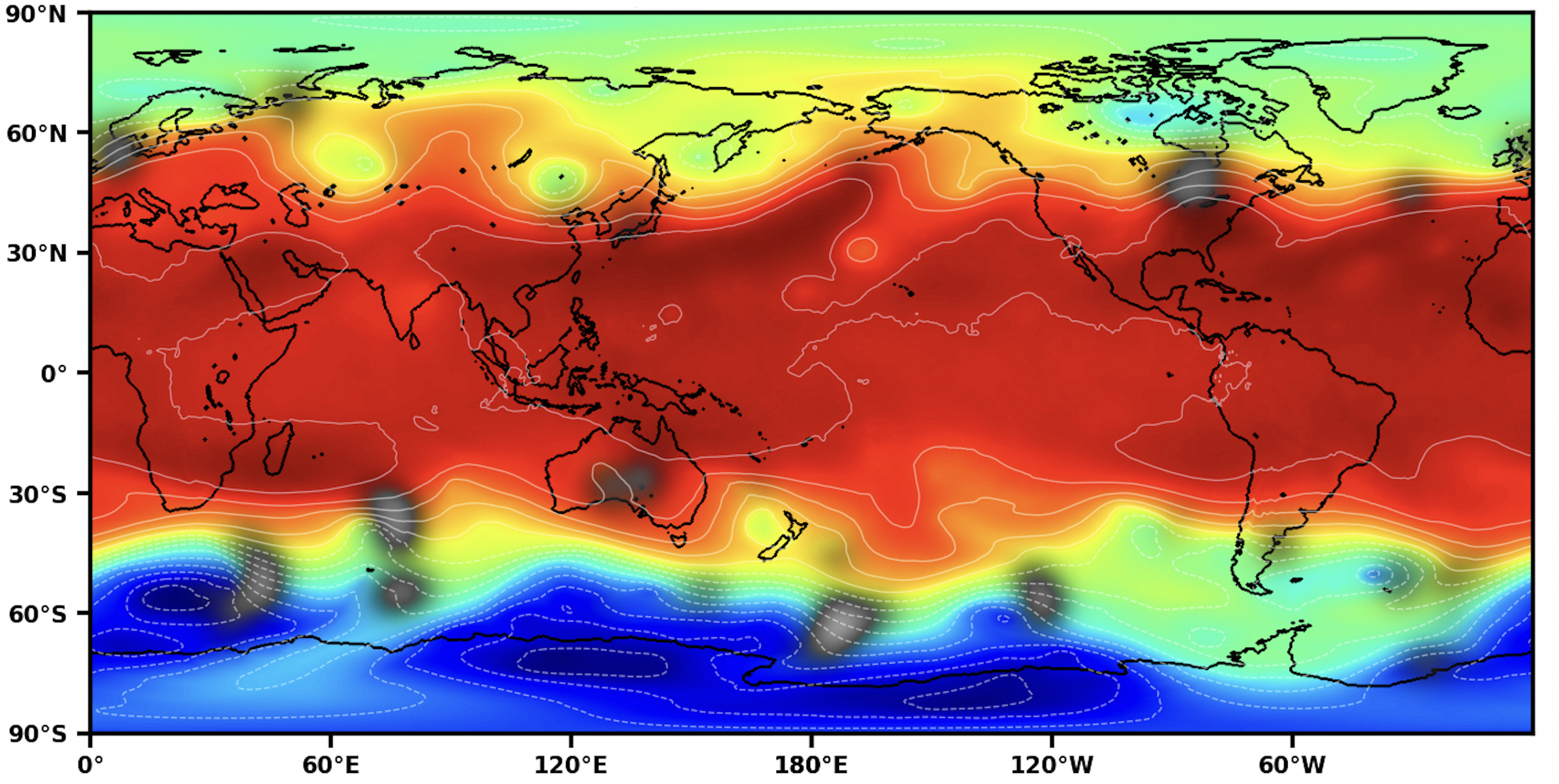} 
    \caption{Geopotential on 500hPa, where red indicates positive values, 
blue indicates negative values, and black mask indicates concept 
activation locations.} 
        \label{Fig:z500}
\end{figure}

\begin{figure}[ht] 
    \centering
    \includegraphics[width=0.5\textwidth]{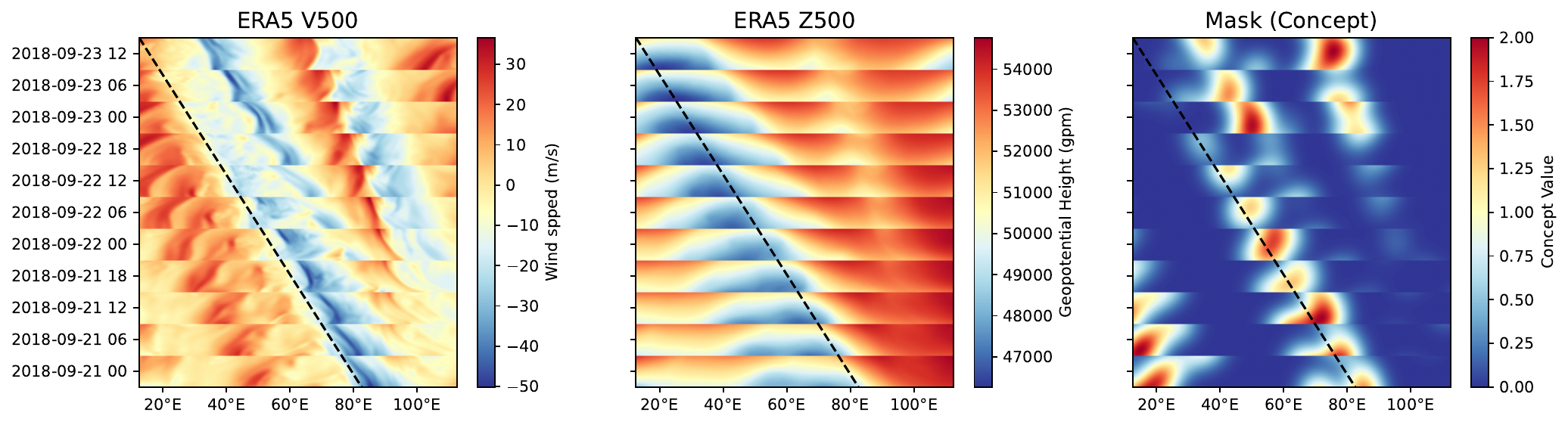} 
    \caption{Co-movement of wind speed maxima, troughs, and the conceptual height center.} 

        \label{Fig:MovementofWave}
\end{figure}

\textbf{Transient  Waves}. Figure~\ref{Fig:v500} and Figure~\ref{Fig:z500} shows the activation map of concept 11736 over the 500 hPa meridional wind field ($v_{500}$, $z_{500}$). Black-shaded regions indicate areas of strong activation. Based on LLM interpretation of the most activated samples, this concept corresponds to atmospheric transient waves: A series of troughs and ridges on quasi-horizontal surfaces in the major belt of upper tropospheric westerlies. which are closely coupled with the mid-latitude upper-level jet stream. These oscillations are not stationary; rather, they continuously evolve and migrate under the influence of the prevailing westerlies. According to the wave characteristic analysis, the extracted transient wave train exhibits distinct zonal periodicity, with a zonal wave number stable between 9 and 10~\ref{Fig:MovementofWave}. These waves propagate in a meandering fashion within the mid-latitude jet stream, constituting the complex and variable synoptic background for mid-to-high latitude regions.




Identification of such wave features is essential for accurate  forecast for  intensity of the tropical cyclone. These transient waves function as a critical determinant in tropical cyclone predictions by perturbing the mid-latitude steering currents. Moreover, given that the trajectory and intensity of tropical cyclones are profoundly sensitive to the surrounding atmospheric state, the dynamical interaction between mid-latitude wave systems and tropical cyclones governs their lifetime. Accurately recording the phase speed and amplitude of these transient waves in forecasting models is necessary to reduce errors in long-term intensity predictions. 




\section{Related Work}

\textbf{Efficient Fine-Tuning}. As the scale of pre-trained models continues to expand, full fine-tuning imposes prohibitive memory and computational costs. This has motivated the development of Parameter-Efficient Fine-Tuning (PEFT)~\cite{ding2023parameter}, which aims to adapt models using a minimal set of trainable parameters while maintaining competitive performance. Early representative methods, such as Adapters~\cite{pfeiffer2020mad, houlsby2019parameter}, insert small feed-forward modules between Transformer layers while freezing the original weights. Although they introduce only a few million parameters, they require structural modifications to the architecture. Alternatively, Prefix-Tuning~\cite{li2021prefix} and Prompt Tuning~\cite{lester2021power} guide model behavior by prepending learnable continuous vectors to input embeddings or hidden states without altering the backbone network. Recently, Low-Rank Adaptation (LoRA) and its variants~\cite{hu2022lora,liu2024dora} have gained widespread adoption due to their simplicity and efficiency. LoRA approximates weight updates through low-rank matrix decomposition: it injects trainable low-rank matrices into attention layers while keeping the original weights frozen. In contrast to these methods, our approach can automatically select concepts for fine-tuning, enabling targeted model adjustments without compromising general capabilities.

\noindent \textbf{Model Interpretability and Intervention}. 
As the scale and diversity of data increase, modifying model behavior through retraining becomes intractable. Consequently, a large of works has shifted its focus to model intervention~\cite{meng2022locating,mengmass,hernandezlinearity}, which refers to the ability to understand, analyze, and modify the internal knowledge and behavior of a model after it has been trained. Model Intervention contains model editing~\cite{de2021editing,dai2022knowledge} and machine unlearning~\cite{yao2024large}. Model editing achieve precise modifications of model behavior by intervening in the model's internal representations or parameters. Machine unlearning aims to remove the influence of specific data from a trained model without the prohibitive cost of retraining from scratch. In contrast, model editing focuses on directly modifying or injecting new knowledge. In the realm of large language models, knowledge editing techniques like ROME~\cite{meng2022locating} have achieved remarkable success, as they can locate and alter the weights associated with specific facts (e.g., "The Eiffel Tower is in Paris"). Furthermore, some studies have attempted to edit more abstract "concepts," for instance, by using linear transformations to suppress or enhance a model's expression of a certain concept~\cite{kim2018interpretability}.

\section{Limitations and Ethical Considerations}
\textbf{Limitations}. Compared to direct fine-tuning, our method requires pre-training an SAE, which introduces a certain training overhead. However, this is a one-time cost, and we argue it is acceptable given the SAE's demonstrated potential to enhance both performance and model interpretability. Additionally, since our method is concept-based, its fine-tuning effectiveness is correlated with concept quality. Fortunately, a large of existing work is dedicated to improving the interpretability of SAEs. Therefore, our method stands to achieve further performance gains by incorporating more advanced SAE techniques in the future.

\noindent\textbf{Ethical Considerations}.
Our study uses meteorological reanalysis/products (e.g., ERA5) and public cyclone tracks. All data are used in aggregated form for forecasting validation and do not include personally identifiable information. No human participants are involved, and no interventions are performed. All data were obtained and used in accordance with the corresponding data-use policies and institutional regulations; thus, IRB review and informed consent are not applicable.

\section{Conclusion} 

The TaCT framework proposed in this study introduces a new paradigm for the application of AI in meteorological science. Its core value can be summarized as follows:

\textbf{Key Contribution to the Domain:} The primary contribution of this work is empowering AI models with a ``surgical'' correction capability. This enables the model to specifically rectify critical failures in predicting high-impact events, such as tropical cyclones, rather than undergoing a broad, general optimization. This directly addresses a core obstacle to the operational trust of AI models: their unreliability in handling high-risk, low-frequency events, thereby paving the way for their deeper integration into disaster mitigation decisions.

\textbf{Specific AI/ML Challenges Tackled:} Through our ``concept-centric'' fine-tuning approach, we effectively address the training challenge of data scarcity for extreme weather. Concurrently, by linking model corrections to physical concepts, we significantly enhance model interpretability, directly confronting the ``black-box'' problem.

\textbf{General Challenges with Using AI/ML:} Our ``concept-gating'' mechanism facilitates precise interventions on specific errors, thereby mitigating the common issue of "catastrophic forgetting" that occurs during specialized optimization. This provides a tangible technical solution for balancing the "specialist" and "generalist" capabilities of AI models.


\section*{GenAI Disclosure}
We used large language models~(LLMs) only to assist with and polish the writing. Specifically, LLMs were used to: (1) check grammar and spelling; and (2) help improve the phrasing in professional academic English. LLMs were not used to generate research ideas, methods, experiments, figures, or results.

\newpage
\bibliographystyle{ACM-Reference-Format}
\bibliography{tact}

\newpage
\appendix
\section{Implementation Details of Sparse Autoencoders}
\label{app:sae_details}
For the SAE, we follow the setting of Gao et al.~\cite{gaoscaling} and employ a Top-k function for sparsity. Additionally, we normalize the hidden embeddings before feeding them into SAE. 
Specifically, given an input climate field $X_t \in \mathbb{R}^{H \times W \times C}$, let the hidden embeddings at the $l$-th layer be $H \in \mathbb{R}^{H \times W \times D}$. 
We first reshape these embeddings into a 2D matrix, which we denote as $H'$, of shape $ (H \times W) \times d$:
\begin{equation}
    H' \in \mathbb{R}^{(H \times W) \times d}.
\end{equation}
We then compute the mean $\mu_H \in \mathbb{R}^{d}$ and standard deviation $\sigma_H \in \mathbb{R}^{d}$ across the second dimension (the flattened spatial dimensions).
Finally, these statistics are used to standardize the matrix $H'$. The standardized matrix, $\hat{H}'$, is computed as follows:
\begin{equation} 
\label{eq:z_norm}
    \hat{H}' = \frac{H' - \mu_{H}}{\sigma_{H} + \epsilon},
\end{equation}
where $\epsilon$ is a small constant added for numerical stability.

To alleviate the issue of ``dead concepts'', we introduce an auxiliary loss, $Aux(\cdot)$, during training. Specifically, for concepts that remain inactive for extended periods, we sample the top-k residuals from the SAE's reconstruction corresponding to these concepts, as formalized in Eq.~\ref{eq:aux} .
\begin{equation}
    \label{eq:aux}
    Aux=||(H'-\hat{H}')- Dec(Topk(z_{dead}))||_{2}.
\end{equation}

\section{Statistical Methods for Concept Discovery}
\label{app:statistical}
As a component of our ablation study (see Section 4.3), we benchmarked our proposed method against a traditional statistical approach for identifying model concepts associated with typhoon events. The objective of this baseline method is to isolate concepts from SAE that exhibit the most significant difference in activation between typhoon and non-typhoon conditions.

The methodology is implemented through the following steps:
\subsection{Data Collection and Patch Labeling}
We first leverage the \textbf{IBTrACS (International Best Track Archive for Climate Stewardship)} dataset to source the historical best-track positions (latitude and longitude) of typhoons at 6-hour intervals. The model's input data, derived from the ERA5 dataset, is segmented into a grid of patches.

Let the set of all input patches be denoted by $\mathcal{P}$. We define a mapping function, $M$, which associates a given typhoon's geo-coordinate with its corresponding patch in $\mathcal{P}$. Using this, we partition all patches into two distinct subsets:

\begin{itemize}
    \item \textbf{$\mathcal{P}_{\text{typhoon}}$}: The set of patches containing a typhoon center.
    \[
    \mathcal{P}_{\text{typhoon}} = \{p \in \mathcal{P} \mid M(\text{lat, lon}) = p, (lat, lon) \in IBTrACS\}
    \]
    \item \textbf{$\mathcal{P}_{\text{non-typhoon}}$}: The set of patches not containing a typhoon center.
    \[
    \mathcal{P}_{\text{non-typhoon}} = \mathcal{P} \setminus \mathcal{P}_{\text{typhoon}}
    \]
\end{itemize}

\subsection{Concept Activation Analysis}

For each concept, $c_i$, within the SAE's concept library, we compute its mean activation value across all patches in the $\mathcal{P}_{\text{typhoon}}$ set. Let $a(c_i, p)$ be the activation of concept $c_i$ for a given patch $p$. The mean activation is formulated as:
\begin{equation}
\bar{A}(c_i, \mathcal{P}_{\text{typhoon}}) = \frac{1}{|\mathcal{P}_{\text{typhoon}}|} \sum_{p \in \mathcal{P}_{\text{typhoon}}} a(c_i, p)
\end{equation}

Similarly, we calculate the mean activation for the same concept across the $\mathcal{P}_{\text{non-typhoon}}$ set:
\begin{equation}
\bar{A}(c_i, \mathcal{P}_{\text{non-typhoon}}) = \frac{1}{|\mathcal{P}_{\text{non-typhoon}}|} \sum_{p \in \mathcal{P}_{\text{non-typhoon}}} a(c_i, p)
\end{equation}

\subsection{Difference Calculation and Ranking}

To quantify the relevance of each concept to the presence of a typhoon, we calculate the delta ($\Delta$) between the two mean activation values:
\begin{equation}
\Delta(c_i) = \bar{A}(c_i, \mathcal{P}_{\text{typhoon}}) - \bar{A}(c_i, \mathcal{P}_{\text{non-typhoon}})
\end{equation}

All SAE concepts are then ranked in descending order based on their resulting $\Delta$ value.

\subsection{Relevant Concept Selection}

Concepts with the highest positive $\Delta$ values are considered to be the most statistically correlated with typhoon phenomena. In our ablation study, these top-ranked concepts were selected to represent the outcome of the statistical discovery method for comparative analysis.

\section{Implementation Details}
\label{app:details}
For our SAE, we targeted the 18th layer of the base model. The SAE was configured with an intermediate dimension of 15,360, a ReLU activation function, and a top-k value of 320. The SAE was trained for 12 hours on 6*A800 GPUs using the AdamW optimizer with a learning rate of 5e-5. 

For LoREFT, we apply interventions to the first 128 token positions.

For counterfactual concept localization, we randomly sampled 438 typhoon data with active tropical cyclone events from the year 2021 as the data for concept localization analysis.

To ensure that the analysis accurately reflects the internal errors of the base model rather than the SAE's own reconstruction errors, the training objective is to keep the output of the model's subsequent layers as consistent as possible, whether the input is the original hidden embedding or the one reconstructed by the SAE, as formalized $\mathcal{F}_{l:L}(h_{l})\approx \mathcal{F}_{l:L}(\hat{h_{l}})$.

\begin{figure*}[ht]
    \centering
    \includegraphics[width=1\textwidth]{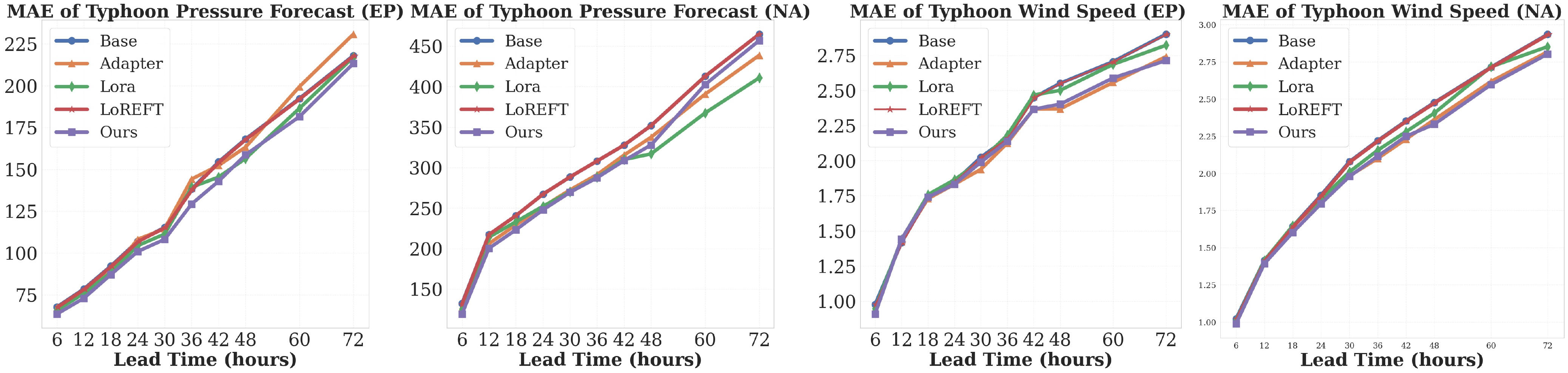} 
    \caption{72-Hours Results. EP denotes the Eastern Pacific and NA denotes the North Atlantic. ``Typhoon pressure'' refers to the minimum sea-level pressure in Pa, and ``Typhoon wind speed'' represents the maximum wind speed in m/s.} 
    \label{fig:other result} 
\end{figure*}

\section{Interpretation Methods}
\label{app:inter}
To automate the process and reduce manual effort, we design an MLLM-based automated interpretation framework. The prompt provided to the MLLM is detailed in Appendix~\ref{app:prompt}. Our input consists of the following components: input-field data, output-field data, the climatological mean, the importance of each variable for the current concept, latitude–longitude information, and land–sea mask information. Specifically, we compute the concept’s input-field data by averaging all variables over a local region centered at each activated location for the same concept. Similarly, we apply the same procedure to obtain the concept’s output-field data. To capture the differences between the concept-related fields and the mean state, we also provide the climatological mean as a reference baseline. We derive the importance of each variable to the concept by computing the gradients of the concept with respect to each variable. For latitude–longitude information, we use binning and take the activation frequency within each bin as the spatial distribution feature. For land–sea information, we categorize locations into three types: land, ocean, and coastline.

\section{Interpretation Prompt}
\label{app:prompt}
\begin{figure*}
    \centering
    \includegraphics[width=1\textwidth]{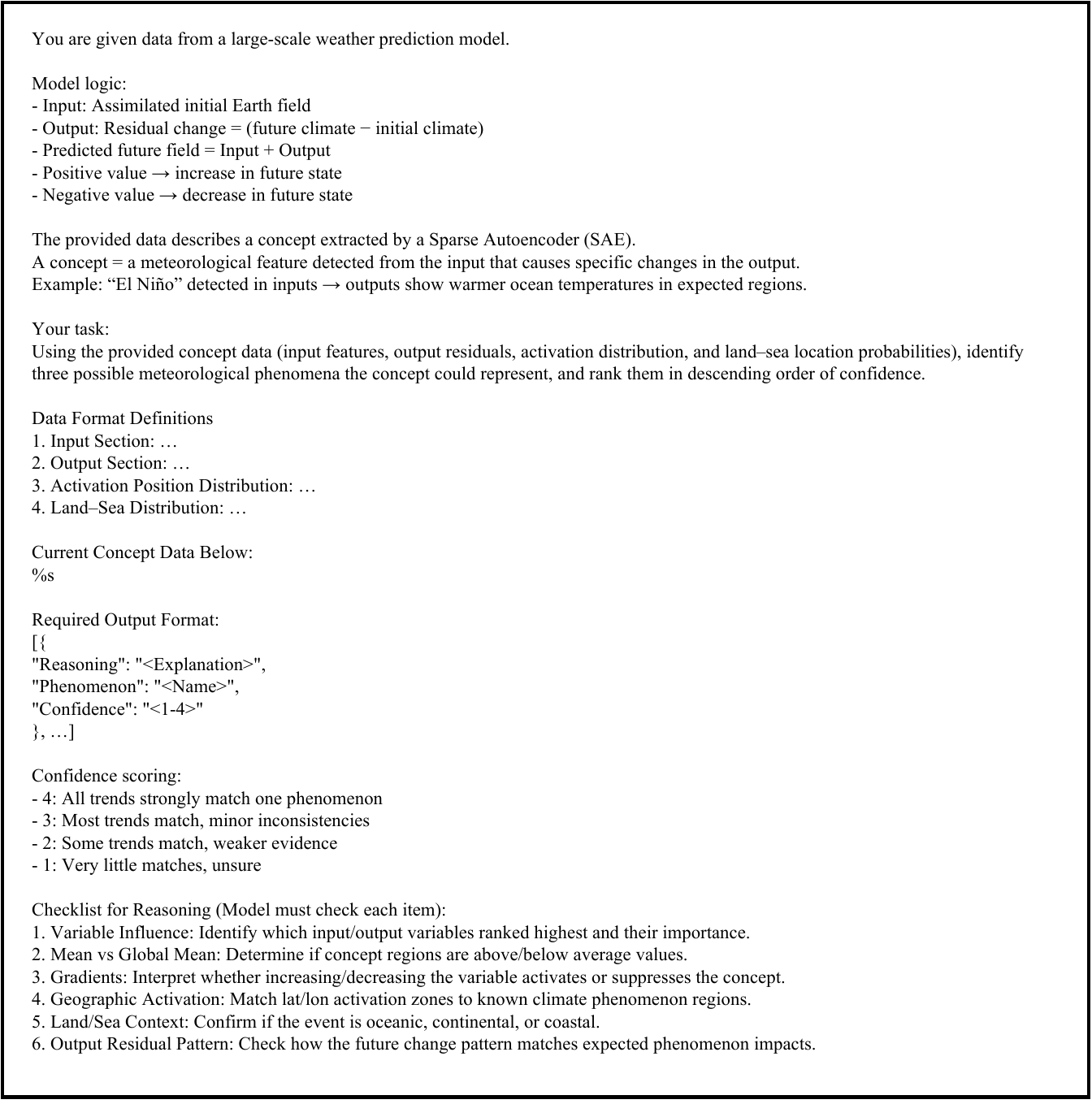} 
    \caption{Prompt for interpreting concepts in multimodal models} 
    \Description{Prompt for interpreting concepts}
    \label{fig:prompt} 
\end{figure*}

The detailed prompt template employed for SAE concept interpretation 
is presented in Figure~\ref{fig:prompt}.

\section{72-Hours Results}
\label{app:72}
Figure~\ref{fig:other result} shows the 72-hour forecast results for other regions, where EP denotes the Eastern Pacific and NA denotes the North Atlantic.

\begin{table}[t]
  \centering
  \caption{Typhoon Data Fine-tuning Results across Different Ocean Basins.}
  \label{tab:typhoon-finetuneing}
  \small
  \setlength{\tabcolsep}{4pt}
  \begin{tabular}{lccccc}
    \toprule
    \textbf{Region} & \textbf{Base Model} & \textbf{LoRA} & \textbf{Adapter} & \textbf{LiREFT} & \textbf{Ours} \\
    \midrule
    North Atlantic   & 132.09 & 124.18 & 123.28 & 131.99 & \textbf{119.21} \\
    Western Pacific  & 82.09  & 78.39  & 78.35  & 82.01  & \textbf{76.88}  \\
    Eastern Pacific  & 67.73  & 65.13  & 64.58  & 67.67  & \textbf{63.53}  \\
    \midrule
    \textbf{Avg}     & 93.97  & 89.23  & 88.74  & 93.89  & \textbf{86.54}  \\
    \textbf{$\Delta$ (\%)} 
                     & 7.91   & 3.00   & 2.48   & 7.83   & -   \\
    \bottomrule
  \end{tabular}
\end{table}

\section{Fine-tuning with Same Data}
\label{app:typhoon}
To compare with models fine-tuned on typhoon data, we evaluate our method against several fine-tuning approaches trained on the same typhoon dataset. Our method still achieves competitive performance. In contrast, other methods may suffer from overfitting due to the scarcity of tropical cyclone data, where excessive fine-tuning on such limited samples can degrade generalization ability, the experimental results are shown in Table~\ref{tab:typhoon-finetuneing}.


\end{document}